# Real-Time Object Detection in Occluded Environment with Background Cluttering Effects Using Deep Learning

Syed Muhammad Aamir*[1], Hongbin Ma *[2], Malak Abid Ali Khan*[3] and Muhammad Aaqib*[4]

*1,2,3 School of Automation, Beijing Institute of Technology, Beijing 100081, China
E-mail: 3820181051@bit.edu.cn
E-mail: mathmhb@bit.edu.cn
E-mail: 3820202079@bit.edu.cn
*4Department of Mechatronics Engineering, University of Engineering and Technology Peshawar, Pakistan
E-mail: enggraaqib@gmail.com

**Abstract.** Detection of small, undetermined moving objects or objects in an occluded environment with a cluttered background is the main problem of computer vision. This greatly affects the detection accuracy of deep learning models. To overcome these problems, we concentrate on deep learning models for real-time detection of cars and tanks in an occluded environment with a cluttered background employing SSD and YOLO algorithms and improved precision of detection and reduce problem facing by these models. The developed method makes the custom dataset and employs a preprocessing technique to clean the noisy dataset. For training the developed model we apply the data augmentation technique to balance and diversify the data. We fine-tuned, trained, and evaluated these models on the established dataset by applying these techniques and highlight the results we get more accurately than without applying these techniques. The accuracy and frame per second of the SSD-Mobilenet v2 model are higher than YOLO V3 and YOLO V4. Furthermore, to employ various techniques like data enhancement, noise reduction, parameter optimization, and model fusion we improve the effectiveness of detection and recognition. We further added counting algorithm, target attributes experimental comparison, and make a graphical user interface system for the developed model with features of object counting, alerts, status, resolution and frame per second. Subsequently, to justify the importance of the developed method analysis of YOLO V3, V4 and SSD were incorporated. Which resulted in the overall completion of the proposed method.

## 1. INTRODUCTION

The techniques employed for identification and locate the object in an image are known as the object detection tasks. Image understanding is a fundamental area of computer vision. Image classification is not enough for understanding images. Object detection is one of the key aspects of image processing and image understanding Precise estimation of the concept and location of the object in an image is called image detection. Object detection and object recognition are two terms that are used interchangeably in computer vision. Object detection is a key research area of computer vision. It attracted many industries such as augmented reality [1] self-driving cars and UAVs, remote sensing [2, 3], ecology investigation [4], and medical science [5]. Datasets that are publicly available and commonly used for object detection training and evaluations include PASCAL VOC, ImageNet, and Microsoft COCO datasets. Many object detectors emerged, and some demonstrate state-of-the-art performance. The key focus of the developed study is deep learning-based object detection. After careful literature review [6] we could broadly divide deep learning-based object detection frameworks into three types which are region proposals (based on region selection) includes R-CNN, Fast R-CNN, and Faster R-CNN, YOLO (You Only Look Once), and SSD (Single Shot MutiBox Detector). R-CNN, Fast R-CNN and Faster R-CNN work in two steps. In the first step they extract the features based on region selection and in the next step, it computes the convolutional neural network and gives the final detection results. Whereas YOLO (You Only Look Once) and SSD (Single Shot Multibox Detector) [7] are single-step models, and they take the input image and compute the convolutional neural network directly in a single step and gives the final detection results. The single-step models like YOLO and SSD are very fast but their accuracy lags when compared to the two-step models like R-CNN, Fast R-CNN and Faster R-CNN. These models give very good accuracy for detection but with less frame per second. Hence, we could not use these models for real-time detection separately. YOLO and SSD work better in real-time detection.

The main findings of this manuscript are as under:
- To categorize the different frameworks in various ways for determining the developed improvements.
- We immense study various hurdles while facing the exact detection of moving objects in an occluded environment.

- We built a graphical user interface to show the live detection of cars and add different features to the system.

This manuscript is organized as follows, section 2 shows the related work followed by the proposed methodology in section 3. Experimentation and results are presented in section 4 and 5, and finally, section 6 conclude this manuscript.

## 2. RELATED WORKS

Before Deep Learning, the task of object detection would be done through various steps, beginning with edge detection and extraction of features from images. Templates were created from these features and then new images were matched with these object templates, usually at multi-scale levels, for the detection of the objects and then localizing these objects present in the image. Histogram of oriented gradients (HOG) [8] is one of the widely used feature extractor techniques in computer vision and image processing applications. The primary concept behind working of the (HOG) descriptor is that the target object's appearance and shape inside an image could be represented by the distribution of intensity gradients and the direction of edges. One of the preliminary neural networks-based object detection algorithms is the sliding window algorithm [9]. In the object detection algorithm, not only want to know whether the object is in the image but also want to know where the object is in the image. The sliding window is one of the earliest object detection algorithms which could detect and locate an object in the image with great accuracy. Algorithms like sliding windows have great detection accuracy but they are slow in the detection of the object. Many applications need real-time object detection,[10] for which the sliding window algorithm is not appropriate. Therefore, a new algorithm called YOLO [11] was introduced in 2016. YOLO (You Only Look Once) is much faster than a sliding window algorithm because YOLO allows the image through the convolutional neural network only once as its name suggests. YOLO V2 [12] is an upgraded version of YOLO V1. YOLO V1 had some limitations, which YOLO V2 tried to improve. The convolutional neural architecture used for feature extraction in YOLO V2 is darknet-19, which is a custom architecture. It uses an additional 11 layers for object detection. First, YOLO V1 could only predict two bounding boxes for each grid cell however, if there are more than two objects whose center lies in the same grid cell, then YOLO will only be able to predict bounding boxes for two objects. YOLO also struggles to detect small bounding boxes in the image as it's a very deep convolutional neural network and can't keep information of small objects because of the max-pooling layers in the network which reduces the receptive field of the network.

Some of the object detectors [13] which take less memory of processing power can't be used in UAVs because their detection accuracy is low and they also fail to detect smaller objects. For this purpose, a new network called Slim-YOLO V3[14] was proposed. It is a variant of YOLO V3. It performs two steps over YOLO V3 to make it lighter and faster. First, it does L1 regularization on the layers to decrease sparsity in the channel. Secondly, it does prunes of less informative feature channels which makes the network lighter and reduces the number of parameters, and also decreases the number of trainable parameters and floating-point operations in the network as compared to the YOLO V3 network. This results in a much lighter network and two times faster than the YOLO V3 [15] network, with a slight decrease in inaccuracy. A single-shot multi-box object detector (SSD) [16] is one of the earliest single shot detectors. Single-shot detection means that the objects were detected and localized on one forward pass. SSD works in two steps. First, it uses a VGG neural network for feature extraction. This creates feature maps for the objects. After creating feature maps, it does convolutional filtering on the feature maps for object detection. A Multi-box detector is a regression technique for creating bounding boxes. SSD is much similar to YOLO algorithms. The difference comes when both predict the bounding box information. YOLO and SSD have different methods of predicting the bounding boxes. Retina-net [17] is one of the most reliable one-step object detection models that operate well with small, dense objects. For this purpose, it has become a popular object detection model for use with aerial and satellite images.

The framework was formed by two developments in existing models of single-stage object detection: characteristic pyramid networks (FPN) and focal loss. Traditionally, in machine vision, custom image pyramids have been used to detect targets at various scales in the image. The pyramid images shown are pyramids of objects built on top of pyramid images. This means that we take the image and divide it into images with lower and lower resolution (thus forming a pyramid). Yilmaz et al. [18] review multiple frameworks for object detection and tracking through a series of videos to classify in various groups. The author recommends multiple problems which are challenging for tracking. The hurdles are sudden motion of the object, object to point camera motion, and occlusion. They also describe the models and features of the best image for object detection.

Zarka et al. [19] developed a method to track the real-time system of human detection and analysis of motion. They suggest a robust framework to track, detect and identify that could be involved with clear effects and occlusion of objects. The author used subtraction of background to achieve the foreground pixels and then the model was applied for object detection and motion analysis.

## 3. PROPOSED METHOD

A method is presented for real-time detection of cars and tanks in an occluded environment with a cluttered background employing SSD and YOLO techniques. The proposed method is fully automatic, real-time and makes incorporates various techniques to achieve the objective of this research. The method performs data construction, the input is then pre-processed and augment data to increase the samples of training. Lastly, we trained YOLO V3 and YOLO V4 [18] models for real-time object detection. The following figure 1 shows the overall process of the proposed system.

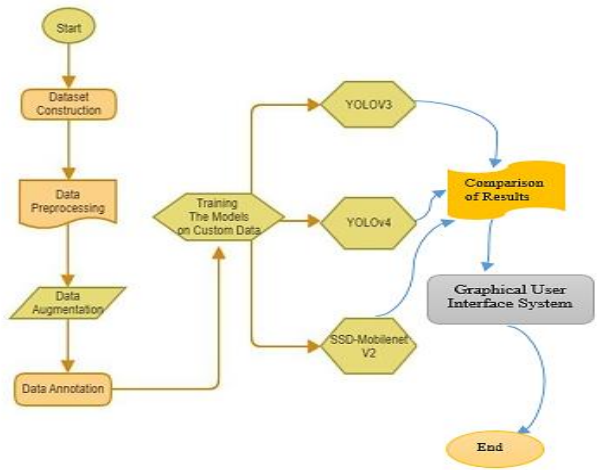

**Fig. 1** Block Diagram of Proposed System

### 3.1. Dataset Construction

Dataset construction is the most challenging part of computer vision. Therefore, to construct a formal dataset, we set a camera far away from the road at an angle that the objects look small, occluded, and with a cluttering effect of the background. We take frames of the video and collect around 60000 images of cars and tanks from those frames. As the camera was fixed at one position and angle and far away from the targeted objects so there was not enough diversity in the dataset. Because all the images are taken from the same angle and without any disturbance. Data preprocessing has a very important and significant influence on the performance of a supervised learning model. To get proper data we employed a technique to remove the noise and enhanced it for model training.

### 3.2. Dataset

We had 2 two types of objects in our image data-set. One type of data set had cars in the images and the other had tanks in the images. These images are the frames extracted from the video sequences. Car's video is taken from the side angle of a traffic signal. Cars in the video are small as the video is recorded from some distance. Similarly, all the cars shown in the images are from the side angle. Therefore, all the training on the cars is done from the side angle. The tank video is taken in an open field. There is only one tank in the video which is roaming around. As the images are taken from the video and mostly they have the same type of cars and tanks. Therefore, in the case of the tank dataset, we have low diversity and had the bounding box information for the objects saved in text file format. For each image, its bounding box information is saved in a separate text file with the same name as the image. Similarly, the bounding box information includes the center point of the height and width of the bounding box then the information is normalized by the height and width of the image. Bounding box normalization is done so that there is not a very big difference in the loss of small and large bounding boxes. The class label for the bounding box was also given in the text file.

The main task for machine learning is labeling the data, therefore, we used a labeling tool to annotate the required objects in the images. After labeling the data we would employ the step of preprocessing i.e., cleaning the data, and then applying data augmentation to increase the dataset and as a result, the model would be trained to achieve better results.

## 4. EXPERIMENTATION

The experimentation phase for object detection in an occluded environment was trained on deep learning models. Due to high accuracy, the SSD-Mobile Net V2 model was chosen for experimentation purposes. In this research work, the model has first fine-tuned three versions of object detection algorithm YOLO V2, YOLO V3, and Slim-YOLO V3 without data augmentation and data preprocessing. During training, the proposed model achieved unsatisfactory results however, the model properly detects, tracks, and recognize the objects in images and videos. The accuracy of YOLO V2, YOLO V3 and Slim-YOLO V3 was 25%, 60.4% and 9.1% respectively. These three object detection frameworks are one of the best object detection algorithms however, it gives unsatisfactory results on the custom dataset due to without employing of data augmentation and preprocessing techniques. While employed the data preprocessing method the proposed model achieved pretty results for the algorithm of YOLO V3, YOLO V4, and SSD-MobileNetv2. At the stages of training, the proposed algorithm passed the images through CNN convolutional neural network and the neural network tries to predict the bounding box information of the images and their loss is calculated with the ground-truth bounding box information.

The following Fig.2 has some samples of images which collect for our custom dataset and annotate these images on labeling software. The objects in our custom dataset are occluded with background cluttering effects. To trained the model for the developed method these custom datasets would be employed.

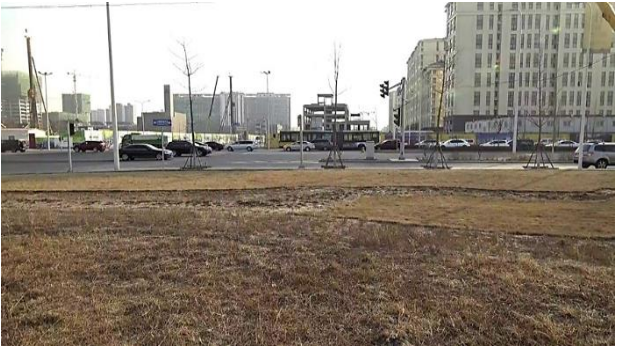

**Fig. 2** Custom Dataset Sample

## 5. RESULTS

A series of experimentation were employed in an occluded environments of object detection with background cluttering effects. In order to exhibit the generalization, this study was implemented employing python with anaconda-based TensorFlow.

The custom dataset was used which was achieved from a video by taking its frames and provided labels to objects is the requirement of the model and then fine-tuned three object detection frameworks and trained them on the custom dataset for two classes i.e., tanks and cars. As the dataset has not enough diversity and the objects were occluded with background cluttering effects in an uncontrolled outdoor environment therefore, some pre-processing and data augmentation steps were done for better results. We used the transfer learning method for SSD-Mobilenetv2 and fine-tuning method for training the different versions of YOLO. These versions of YOLO and SSD are among the best object detection algorithms. Then trained YOLO V2, YOLO V3, and Slim-YOLO V3 models on the custom dataset without preprocessing and data augmentation but without applying these techniques we achieved unsatisfactory results. In this work, we evaluated and compared the results of these two versions of YOLO and SSD-Mobilenet-v2 on the dataset and detect cars and tanks in real-time in an occluded environment with background cluttering effects.

### 5.1. Implementing YOLO V3

We utilized the darknet implementation for fine-tuning YOLO V3 on the custom dataset. The algorithm can detects objects in images at three stages to detect small, medium, and large objects. Then modified the configuration file of YOLO V3 to detect two classes, set the learning rate to 0.001 for the stability of losses and keep the batch size of 8 so that batch will have enough examples to get different features of the images. Similarly, tested the model for detections after every 100 epochs and saved the best model. The following fig.3 has the detection, evaluation, and graphical results of YOLO V3 trained on the custom dataset.

As presented in Fig.3 the error training loss and evaluation graph of YOLO V3 and the number of training epochs was kept at 20k with an initial learning rate (LR) of 0.001. The LR has been approaching 0.1 times of initial LR at 16k and 18k epochs respectively.

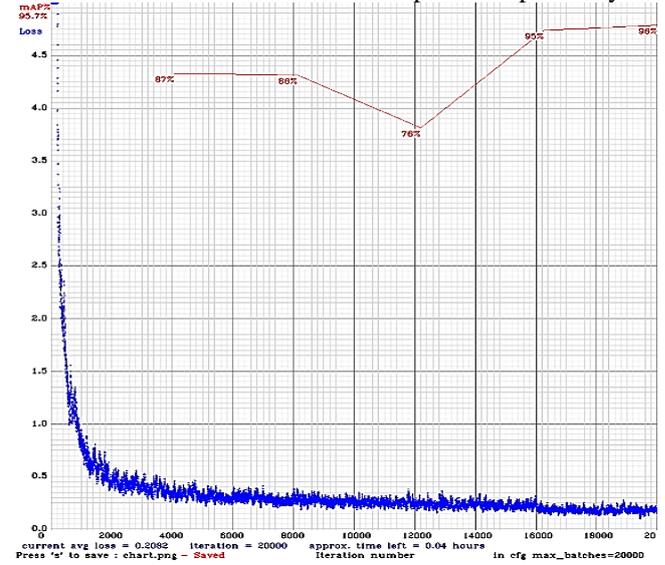

**Fig. 3** Training Graph of YOLO V3

The following Fig.4 are the real-time detection results of the model YOLO V3 in an occluded environment with background cluttering effect which trained on the custom dataset. For training the model we employ preprocessing to clean the dataset and then data augmentation to produce diversity in the dataset and after that fine-tune the model according to respective classes and get pretty results while testing the data. To test the capabilities of the proposed model the mean average precision (mAP) of 82.6% was achieved.

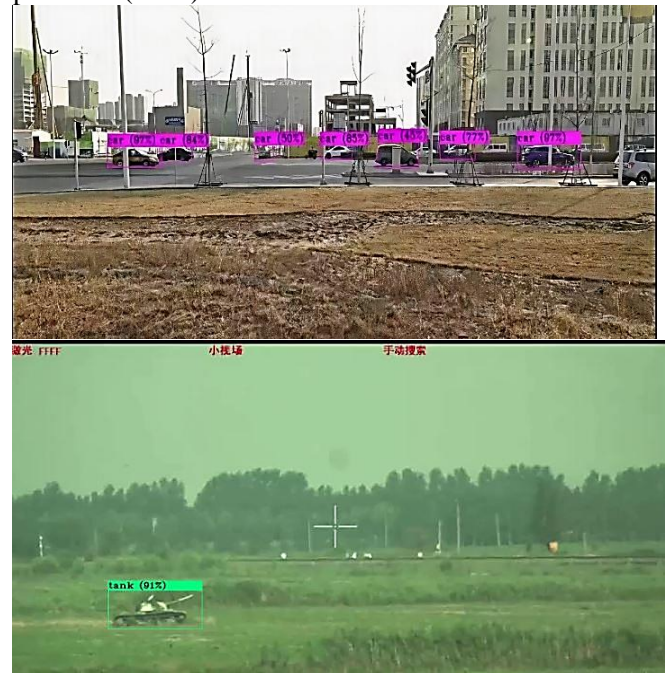

**Fig. 4** Real-Time Detection in Occluded Scenario and Cluttered Background

In order to justify the effectiveness of the proposed experimentation a comparative analysis was taken into account without employing preprocessing and data augmentation and it represents the mAp of YOLO V2, YOLO V3, and Slim-YOLO V3 results in fig.5.

Similarly, the experimentation was done with the help of preprocessing to remove the unwanted data and data augmentation for generating diversity in the dataset. Subsequently, the proposed model was again trained on YOLO and SSD-Mobilenet-V2 and achieved more accurate and better results than the models applied without preprocessing and data augmentation techniques.

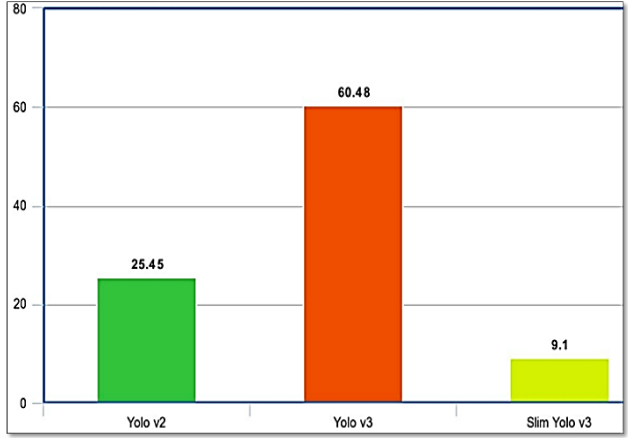

**Fig. 5** Map of YOLO Without Preprocessing and Augmentation

During training, the model of YOLO V2 on a custom dataset without employing preprocessing could see that some of the cars are detected with a low score while some cars are detected with good scores. Therefore, for real-time detection, the overall results are not satisfactory as depicted in Fig.6. Here as shown in the following figure that the black car is occluded and some portion of the car is covered by the tree which is in front of it but still it is detected by the model with a pretty score but the cars which are in white are not detected with good score.

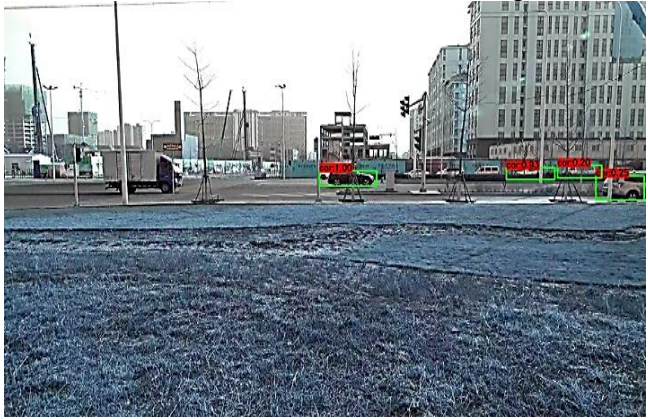

**Fig. 6** Detection Results of YOLO V2 Without Preprocessing and Augmentation

Similarly, in Fig.7 the detection results of the YOLO V3 model in which black car is occluded and some part of the car is covered by the white car and some part is covered by the tree and wall and it's not detected successfully but the white car is detected with a low score. While in the condition of the tank, it is also detected but the detection rate is low which is not acceptable in the case of real-time detection.

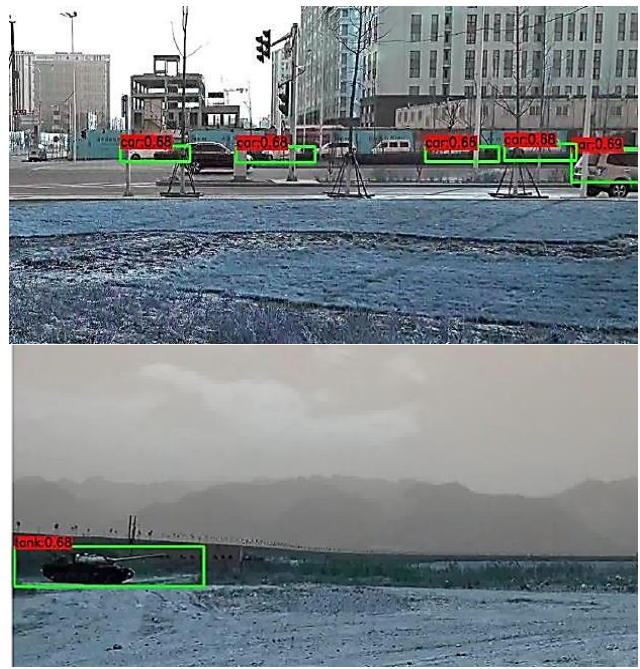

**Fig. 7** YOLOV3 Detection Results Without Preprocessing and Augmentation

During training on a custom dataset, only one car was detected employing Slim-YOLO V3 algorithm as depicted in Fig.8. The results of Slim-YOLO V3 are unsatisfactory for real-time detection. Here the following image black car is occluded however, it's not identified by the Slim-YOLO V3 model. The developed method concluded from the results that the model was employed and fined tuned and trained on the dataset to get average results.

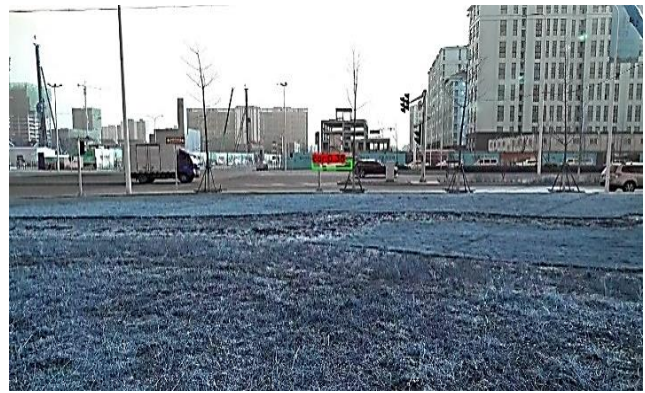

**Fig. 8** Detection Results of SLIM YOLO V3 Without Preprocessing and Augmentation

### 5.2. YOLO V4 Results

We employed the YOLO V4 model and trained with the same learning rate. However, network response was different because of its core structures are different. We used CSPDarknet53 as a backbone, SPP, and PAN models as neck and YOLO-v3 as ahead. LR has been reduced to .1 times at 16k and 18k. After 18k iterations, the model accuracy is not improving further.

Fig.9 shows that the tanks are classified by the model with a good score. As the tanks have the same augmentation can detect the tanks even if it has background cluttering effect color with the background and it's very difficult for a model to detect objects in such scenario but the model which we fine-tuned and trained on our custom dataset after preprocessing and data augmentation get satisfactory results.

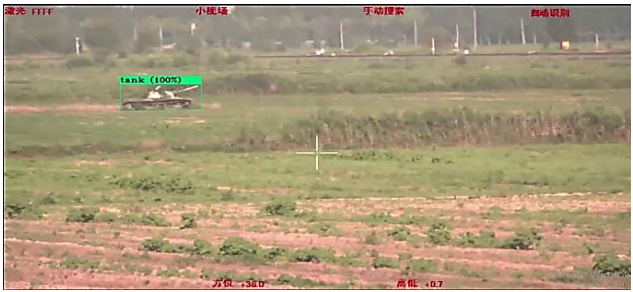

**Fig. 9** Real-Time Detection of Tanks with Background Cluttering Effects

If two objects have the same color and merged or close to each other then it's more difficult for a model to detect and recognized the objects in such a condition therefore, the proposed model would detect and classify the data with good scores in an occluded environment. It is depicted in fig. 10 that some cars are very closed with each other and a lot of factors have been seemed in front of the car and covered some portion of the car but still, it detected and classified the car accurately.

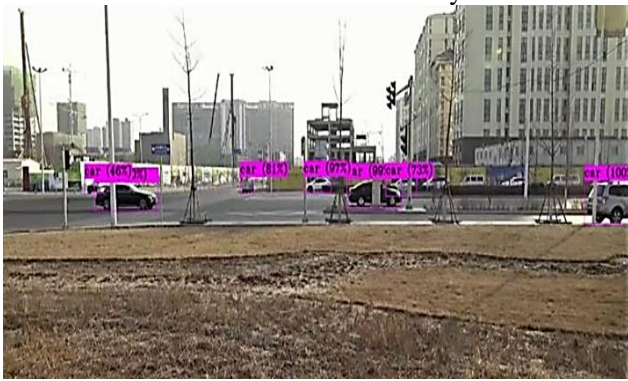

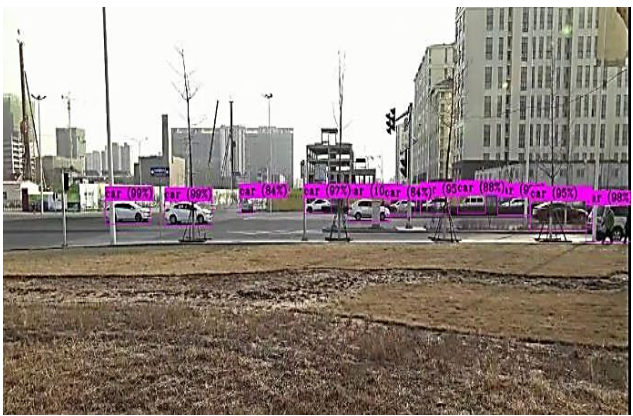

**Fig. 10** Real-Time Detection of Cars in Occluded Environment

Further computations were performed to get the values from the confusion matrix and the threshold value of 0.25 true positive is 1734 and false negative is 201, the average intersection over union is 67.21%. Similarly, the intersection over the union threshold is 75% used under the curve for each unique recall. Mean average precision (mAp @ 0.75) is equal to 0.823662. After the evaluation of the SSD-Mobilenet v2 model, the proposed technique achieved 73.92% average precision of cars at a value of 50 for intersection over union threshold and 100% average precision for tanks at a threshold value of 50 IOU. The mean average precision we get at 0.5 IOU (intersection over union) is 0.866121 which is equal to 86.6%. After evaluation of the model, the classification loss and localization loss were 0.79 and 0.39 respectively.

Here in the following fig. 11 are the detection results of our model which we trained on our custom dataset of cars and tanks after preprocessing and data augmentation for better detection results Here we could notice the cars are parked very near with each other and even on the naked eye we can't see the full image of every car, as some parts of the cars can be seen but still the model detects the cars more efficiently with better accuracy rate. The cars are merged and even some cars have the same white color and the tank has the same color with the background which is yellow and the model can detect cars and tanks in such environments.

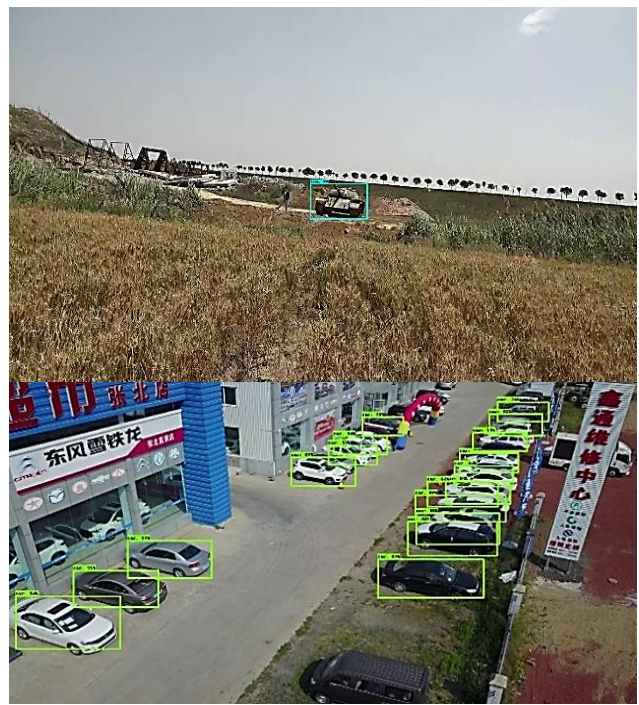

**Fig. 11** Detection Result of SSD-MOBILE NET-V2 in Occluded Environment

It is noticed that the real-time detection results of unchanged existing techniques have low scores and not classified the data correctly, even they are not fully or partially occluded. While the detection results of improved models can successfully detect objects in an occluded environment in real-time. So it is depicted in fig.11 that some part of the white car is covered by the black car which is known as occluded but still the algorithm is efficient to detect all the vehicles successfully. We get these results after applying to preprocess and data augmentation techniques. We test our models with both custom and standard datasets. After applying data augmentation and preprocessing we fine-tune the algorithm and set the best parameters for training the model and also the addition of extra features to the algorithm like counting, class id, recording, database, and status. The cars in the following figure are mostly covered by trees, traffic light poles, walls, and some parts of cars are covered by the same color cars which are known as occluded vehicles. Hence the proposed algorithm detects each vehicle successfully and it proves that the model was efficient in real-time with good prediction scores.

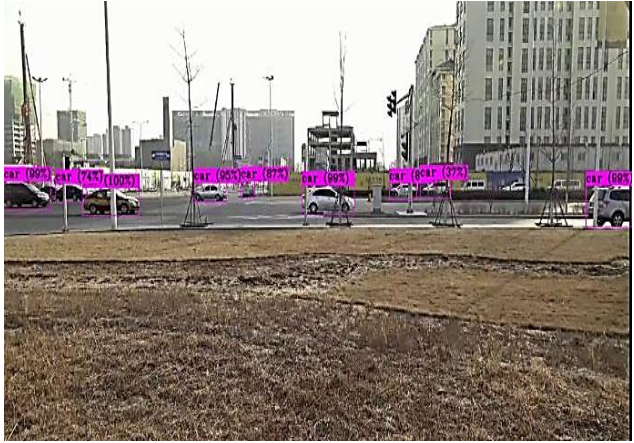

**Fig. 12** Results of Improved Algorithm

A summary of multiple models is shown in the following table. Different networks and frameworks were utilized to get better rates of performance. SSD, YOLO V4, and YOLO V3 have 86%, 82% and 78% mAp respectively.

Table 1 Different models result from summary

| Network | Map | FPS | Network |
|---|---|---|---|
| SSD–Mobile net V2 | 86.6% | 104 | SSD – Mobile net V2 |
| YOLO V4 | 82.2 | 58 | YOLO V4 |
| YOLO V3 | 78 | 63 | YOLO V3 |

### 5.3. GUI RESULTS

It is depicted in Fig. 13, that the graphical user interface shows the live detection of cars because of adding different functionalities and features to the system such as it shows the frame per second of the video as well as the resolution and give us the counting results which is one of the best parts of experimentation. It also represents the states where it's tracking or not and gives us the alert based on counting. It has different controls like start, stop, start recording, and stop recording.

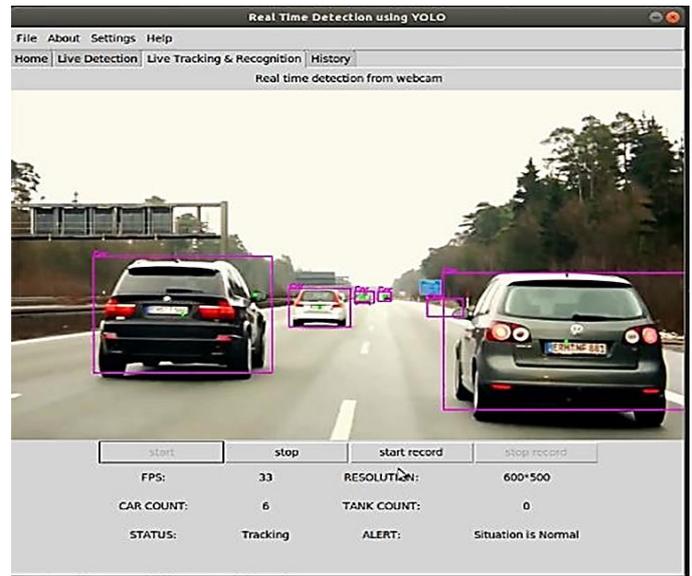

**Fig. 13** Live Detection Results on GUI

The detection results of tanks on the graphical user interface are shown in Fig. 14. There is only one tank that exists in the video so the count status is 1. As for the current video, the frame per second is 34 and it depends upon the system architecture that the model employed. If we use a good system with a good GPU for testing the model, then it will be greater in number. As it can be used in the military sector for the detection, recognition, tracking, and counting of tanks in the war field as well in the army bases.

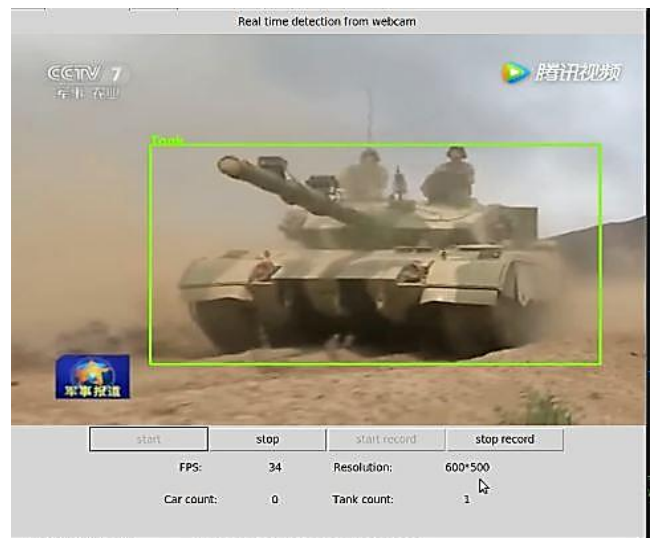

**Fig. 14** Detection Result of Tank on GUI

It could also be used in the transportation sector like parking slots, parking buildings in the airports and different kinds of the bus stand as well as entering points of any place where we need counting and want to know the number of cars or tanks inside that place.

### 6. CONCLUSION AND FUTURE WORK

This study involved multiple algorithms and hence fine-tuned YOLO V2, YOLO V3, and Slim-YOLO V3

on a custom dataset without employing the technique of data preprocessing and data augmentation and achieved unsatisfactory results. Similarly, performing of data augmentation and preprocessing techniques and fine-tune YOLO V3, YOLO V4 models, additionally used the method of transfer learning for Mobilenet-SSD and trained these models on custom dataset to achieved good accuracy with satisfactory frame per seconds for real-time detection. We diversify the dataset for better training to get good accuracy of detection while the objects are small, occluded, and also with background cluttering effect. After applying the proposed model, we get 78% mAp with 58 FPS on the YOLO V3 model, 82.3 % mAp with 63 FPS on the YOLO V4 model, and 86.6% mAp with 104 FPS on SSD-Mobilenet V2. Therefore, the developed SSD-Mobilenet v2 is more accurate and robust in this sceniorio. The training was performed on different configurations of YOLO and was tested on the test split of the data, then the results were compared with the proposed and already established method. Among these models, SSD-Mobilenet v2 gives better accuracy with more robustness, which is 86.6 mAp with 104 frames per second. For real time detection and tracking we build graphical user interface system which shows different functionalities and features like real-time detection, real-time tracking and recognition, car count, tanks count, alerts, frame per second, resolution, status and record.

In the future, we would focus on developing the developed algorithm for faster, accurate, and robust detection. We would be able to change the base of the model to develop a new model according to our need and could add more features to the graphical user interface system which would be used in more applications to solve real-world problems.